\documentclass[10pt,twocolumn,letterpaper]{article}
\usepackage{iccv}
\usepackage{times}
\usepackage{epsfig}
\usepackage{graphicx}
\usepackage{amsmath}
\usepackage{amssymb}
\usepackage{subcaption}
\usepackage{epstopdf}
\usepackage{cases}
\usepackage{multirow}
\usepackage{booktabs}
\usepackage{caption}
\usepackage{array}
\usepackage{makecell}
\usepackage{ragged2e}   
\usepackage{siunitx}    

\usepackage[pagebackref=true,breaklinks=true,letterpaper=true,colorlinks,bookmarks=false]{hyperref}

\iccvfinalcopy 

\def\iccvPaperID{3135} 

\ificcvfinal\fi

\begin{document}

\title{Both Spatial and Frequency Cues Contribute to High-Fidelity Image Inpainting}

\author{Ze Lu\thanks{indicates equal contributions.},  Yalei Lv$^*$,  Wenqi Wang, Pengfei Xiong\thanks{corresponding author. xiongpengfei2019@gmail.com} \\
Shopee MMU}

\maketitle
\ificcvfinal\thispagestyle{empty}\fi

\begin{abstract}
Deep generative approaches have obtained great success in image inpainting recently.
However, most generative inpainting networks suffer from either over-smooth results or aliasing artifacts. The former lacks high-frequency details, while the latter lacks semantic structure.
To address this issue, we propose an effective Frequency-Spatial Complementary Network (FSCN) by exploiting rich semantic information in both spatial and frequency domains.
Specifically, we introduce an extra Frequency Branch and Frequency Loss on the spatial-based network to impose direct supervision on the frequency information, and propose a Frequency-Spatial Cross-Attention Block (FSCAB) to fuse multi-domain features and combine the corresponding characteristics.
With our FSCAB, the inpainting network is capable of capturing frequency information and preserving visual consistency simultaneously.
Extensive quantitative and qualitative experiments demonstrate that our inpainting network can effectively achieve superior results, outperforming previous state-of-the-art approaches with significantly fewer parameters and less computation cost. The code will be released soon.
\end{abstract}


\section{Introduction}

\begin{figure}[t!]
\centering
\includegraphics[width=1.02\columnwidth]{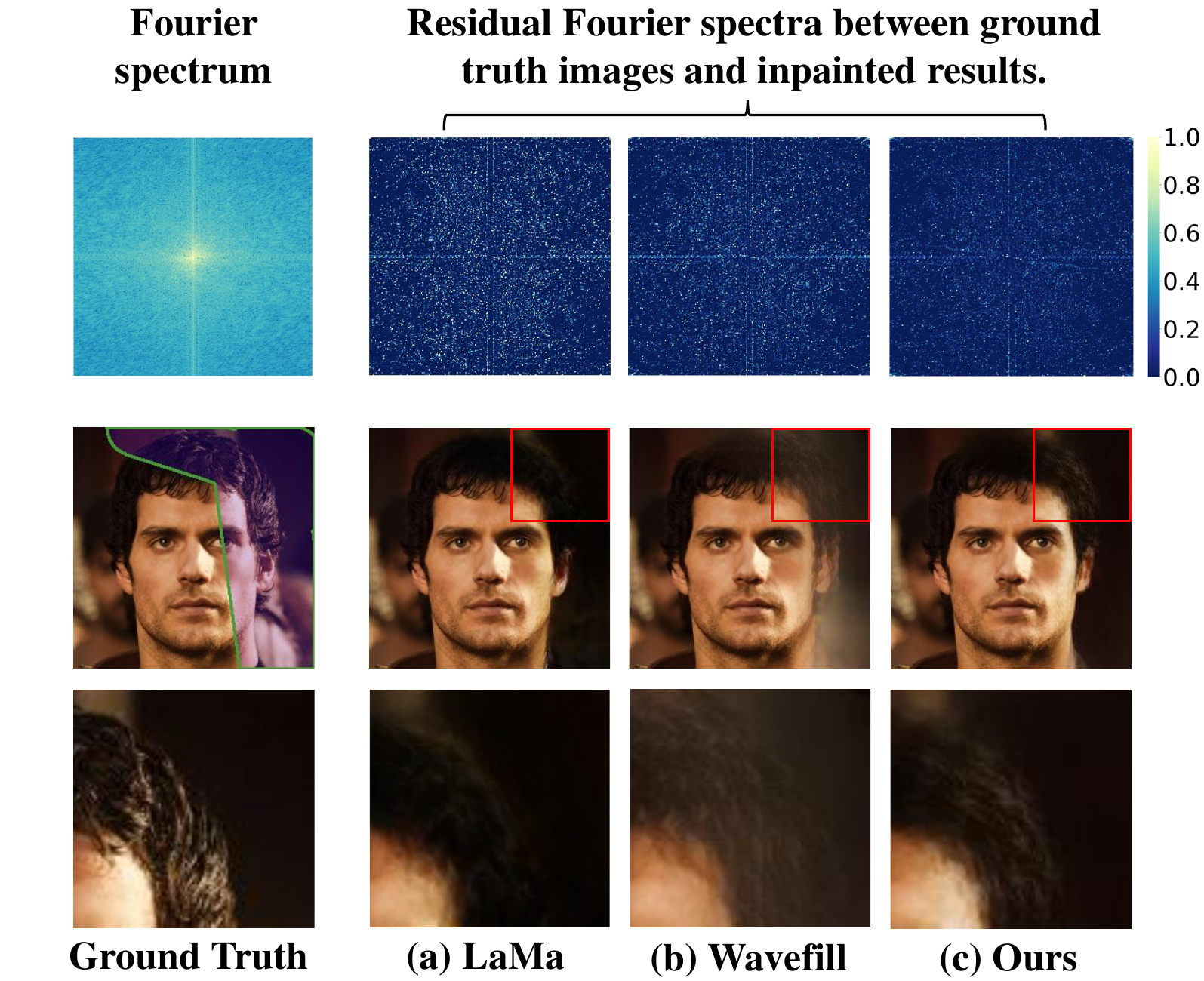} 
\caption{ We show an image and the corresponding frequency spectrum in each column. Zoom in for a better view.
The center point of the frequency spectrum denotes the lowest frequency and the outermost region denotes the highest frequency.
LaMa~\cite{lama} loses much high-frequency information and produces over-smooth artifacts in the hair area. Wavefill~\cite{wavefill} neglects to explore the spatial features and produces unpleasant perceptual quality, e.g. aliasing artifacts.
Compared with them, our result is more similar to the ground truth in both the frequency and spatial domains.} 
\vspace{-4mm}
\label{fig:intro}
\end{figure}

\noindent 
Image inpainting aims to recover visually realistic texture in incomplete images.
It has played an important role in various applications, such as object removal, old photo restoration, and face completion. 
Recently, significant progress has been achieved with the development of various powerful deep learning based methods
~\cite{schwarz2021frequency,durall2020watch,jiang2021focal,glama2022,partial2018,multicolumn2018,zoomtoinpaint2022,Huang2022,lama,wavefill,li2022mat}.
However, restoring realistic and high-fidelity images still remains a challenging task.

Most of the existing inpainting methods ~\cite{lama,partialConv,multicolumn2018,deepfillv2,edgeconnect} explore only spatial features, with no exploitation of the frequency features. These methods can usually recover plausible global textures. However, they typically generate frequency spectra deviating from the ground-truth spectra, leading to unsatisfied perceptual quality in the visual space, such as over-smooth or checkerboard artifacts. As shown in Fig.~\ref{fig:intro}, LaMa~\cite{lama}, one of the state-of-the-art spatial domain methods, suffers from the loss of high-frequency information, which corresponds to a lack of important fine-grained details in the spatial space. 
The frequency gap may be attributed to the inherent bias of neural networks. 
Xu et al. \cite{xu2019training} propose F-Principle that neural networks first quickly capture low-frequency components and then slowly capture the high-frequency ones. Therefore, with no supervision on the high-frequency components, neural networks fail to maintain important high-frequency information.



In order to decrease the deviation in the frequency space, frequency-based methods, such as Wavefill~\cite{wavefill}, DFT~\cite{dft2020}, GLama~\cite{glama2022}, decompose the input image into multiple frequency bands using discrete wavelet transform and recover the corrupted regions in each 
band individually and explicitly.
The separately completed features in different frequency bands are then transformed back into the spatial domain to produce the final image.
Nevertheless, the final image generated by directly combining the individually completed frequency features suffers from neglecting features in the space domain and incorrect structures or distortions, resulting in poor perceptual quality as shown in Fig.~\ref{fig:intro}.

To address the above problems, we propose a Frequency-Spatial Complementary Network (FSCN), which exploits both the frequency and spatial information to improve image inpainting quality. 
The key idea of FSCN is to introduce an extra Frequency Branch and an additional Frequency Loss on the spatial-based network to explicitly provide a frequency constraint, and a Frequency-Spatial Cross-Attention Block (FSCAB) to fuse multi-domain features in both the spatial and frequency branches.

Specifically, in the frequency branch, we apply the Fast Fourier Transformation to each image and adopt several residual blocks to fit training data in the frequency domain.
The frequency branch provides great impedance to prevent the deviation of important frequency components due to the inherent bias of neural networks.
Existing inpainting methods usually adopt loss functions in the spatial domain, while we introduce an additional frequency loss.
The frequency loss directly regulates the consistency and penalizes deviation of images in the frequency domain during training.
After that, a fusion block is adopted to enhance mutual information in frequency and spatial domains and explicitly supervise each other via a cross-attention block (FSCAB). By rescaling original spatial features based on the correlation with frequency features, FSCAB can fully combine the advantages of both frequency and spatial features, not only to capture the frequency information but also to achieve visually plausible results.

We conduct extensive experiments on the CelebA-HQ~\cite{celeba-hq} and Places~\cite{place} datasets for evaluation.
Quantitative and qualitative results demonstrate that FSCN can effectively achieve superior results. 
Furthermore, our model outperforms MAT~\cite{li2022mat} in terms of the SSIM metric ($0.9184$ versus $0.9033$), with parameters decreasing from $61.6$M to $12.3$M and MACs decreasing from $418.7$G to $62.72$G.

The main contributions are summarized as follows:
\begin{itemize}
\item We propose a Frequency-Spatial Complementary Network, which includes an extra Frequency Branch and a Frequency-Spatial Cross-Attention Block to fuse multi-domain features, taking full advantage of characteristics of both the frequency and spatial features. To the best of our knowledge, this is the first demonstration of joint frequency and spatial domains in feature constraint.
\item We propose an additional Frequency Loss with regard to the commonly used spatial-domain network. Such design helps narrow down the frequency gap due to the inherent bias of neural networks and ensures that the inpainted images share similar frequency spectra with ground truth images.
\item We demonstrate that our proposed FSCN can effectively improve model performance against existing SOTA methods in terms of both quantitative and qualitative metrics by extensive experimental studies on public datasets. It brings visually more realistic results with significantly less computational cost.
\end{itemize}

\begin{figure*}[t!]
\centering
\includegraphics[width=\textwidth]{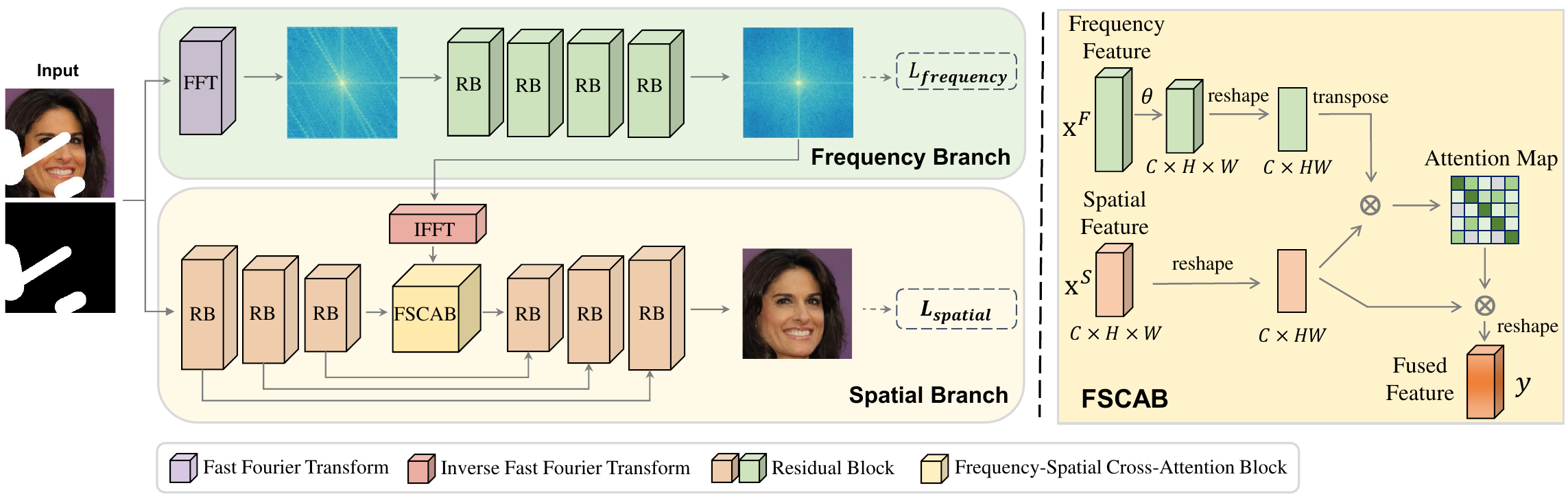} 
\caption{The overall architecture of our proposed FSCN. An input image is fed into the spatial branch and frequency branch simultaneously, and the generated features are then fused via a Frequency-Spatial Cross-Attention Block (FSCAB) to combine the characteristics of both the frequency and spatial features. The frequency loss and spatial loss are introduced to regulate both frequency and spatial consistency for high-fidelity reconstruction.} 
\vspace{-4mm}
\label{fig:archi}
\end{figure*}

\section{Related Works}
\subsection{Deep Image Inpainting}
Recently, deep learning based image inpainting algorithms show superior performance over the traditional ones \cite{liu2018structure,Nguyen2019new,sun2022learning,zhang2022wnet,sun2022TSINIT,deepfillv1}.
Among them, \cite{pathak2016context} first proposes a deep learning inpainting method with an encoder-decoder architecture.
Partial convolution~\cite{partialConv} and gated convolution~\cite{deepfillv2} are proposed where the convolution is masked and conditioned only on valid pixels.
In \cite{wu2022deep}, a two-stage generative model is proposed to handle the image inpainting task in an end-to-end manner. 
Edgeconnect~\cite{edgeconnect} is proposed to predict the complete edge map in the first stage, which is used as prior knowledge to guide the second-stage inpainting network to restore more refined results.
Transformers and convolutions are also unified to model long-range interactions to guarantee high-fidelity image inpainting \cite{li2022mat}. Despite their immense success, most methods only explore spatial-based features and pose no constraint on the high-frequency components.
Due to the inherent bias of neural networks, they will first quickly capture low-frequency components and then slowly capture the high-frequency ones, leading to the lack of important high-frequency details, which can be perceived as over-smooth artifacts. 
Our method introduces an extra frequency branch and an additional frequency loss to the commonly used spatial-domain network to narrow the gap between the inpainted image and the ground truth image in the frequency domain.

\subsection{Frequency-based Methods}
Frequency analysis has been incorporated into various computer vision tasks, such as quality enhancement~\cite{wang2020multi, zhang2022wavelet}, image editing~\cite{gao2021high} and image super-resolution~\cite{deng2019wavelet}.
Xu et al. \cite{xu2020learning} show that frequency-domain learning can better preserve image information than the common spatial approaches and consequently achieve better performance.
A focal frequency loss \cite{jiang2021focal} is introduced to allow neural networks to adaptively focus on the frequency components that are more difficult to generate by mining hard frequencies.
Since the reconstruction loss and adversarial loss focus on generating different frequency components, WaveFill \cite{wavefill} is proposed to decompose images into multiple frequency bands by discrete wavelet transform to mitigate inter-frequency conflicts. 
To tackle the problem of blind face inpainting, a two-stage method named FT-TDR is proposed in \cite{wang2022fttdr} with a network detecting the corrupted regions with the exploration of frequency information. 
These frequency-based methods try to narrow the information gap in the frequency domain, while they still obtain unpleasant perceptual quality, as they fail to fully exploit spatial-domain features. Different from them, we propose a Frequency-Spatial Cross-Attention Block to fuse the cross-domain features and make full advantage of the feature correlations both in the frequency and spatial domains, achieving both less frequency loss and better perceptible results.

\section{Proposed Method}
\vspace{-1mm}

The overall framework of our Frequency-Spatial Complementar Network (FSCN) is shown in Fig.~\ref{fig:archi}.
Specifically, we adopt the frequency and spatial branches simultaneously to fill the corrupted regions from respective perspectives and regulate both the frequency and spatial consistency.
Then, we introduce the Frequency-Spatial Cross-Attention Block (FSCAB) to fuse multi-domain features and make full use of the advantages of these branches.
In the following, we describe our proposed method along with the training objective functions in detail.

\subsection{Frequency Branch}
According to the F-Principle \cite{xu2019training}, with no constraint on the frequency components, neural networks tend to fit them from low to high frequencies. 
As a result, the generated frequency spectra often deviate from the ground truth, which corresponds to different artifacts in the visual space, e.g., over-smooth.
To this end, we introduce a frequency branch to complete the input image from the frequency perspective and regulate frequency consistency, which helps to recover more high-frequency details and improve image quality.

We use the Fast Fourier Transformation (FFT) to convert an image into its frequency representation.
Given a real 2D image $I$ of size $H \times W$, we apply the Discrete Fourier Transform $\mathcal{F}$:
\begin{equation}
    \mathcal{F}(I)(u,v) = \sum^{H}_{h=1} \sum^{W}_{w=1} I(h,w) \cdot e^{-2\pi i \left( \frac{hu}{H} + \frac{wv}{W} \right) },
\end{equation}
where $I(h,w)$ is the pixel value coordinated at  $(h,w)$ in the spatial domain, $\mathcal{F}(I)(u,v)$ is the complex frequency value  coordinated at  $(u,v)$ in the frequency domain, and $i$ is the imaginary unit.

After transforming an image into the frequency space, we apply several residual blocks to extract features and recover corrupted regions from the frequency perspective.
The generated frequency features are then fused with spatial features. 
According to the spectral convolution theorem in Fourier theory, each point in the spectral domain would globally affect all the input pixels of the spatial domain, which contributes to non-local receptive fields and global feature representational ability.

\subsection{Spatial Branch}
Complement to the frequency branch that helps recover high-frequency details, the spatial branch is introduced to explore positional and structural information, which is beneficial for reducing artifacts such as misalignment and aliasing. In the spatial branch, we design a UNet architecture~\cite{unet} based on residual blocks.
Skip connections are added to an encoder-decoder to connect feature maps and masks.
The input to the final residual block includes the concatenation of the original input image with holes and original mask, encouraging the model to duplicate non-hole pixels. 
The spatial features are extracted via convolutions, which only operate in a local manner and interact with adjacent pixels to encourage local consistency. 
Therefore, the frequency branch and the spatial branch are complementary in extracting both local and global features.

\subsection{Frequency-Spatial Cross-Attention Block}
Since the frequency branch and the spatial branch complementarily address different problems, a Frequency-Spatial Cross-Attention Block (FSCAB) is added to the bottleneck of the UNet architecture to aggregate the frequency and spatial features.

As illustrated in Fig.~\ref{fig:archi}, we propose FSCAB inspired by the principle of non-local block~\cite{wang2018non}.
Given the input frequency feature $x^F$ and the input spatial feature $x^S$, we first implement a convolution $\theta$ to transform the frequency feature $x^F$ to an embedding space.
Then we reshape $x^F$ into $C \times HW$ and $x^S$ into $C \times HW$, where $H$ and $W$ denote the height and width respectively, and $C$ denotes the number of channels. 

Afterwards, we conduct a matrix multiplication between the transpose of frequency feature and spatial feature to calculate the inter-pixel attention scores.
The attention scores are activated by ReLU to suppress negative values,
\begin{equation}
    f(x_i^F,x_j^S) = {\rm ReLU} \left(\frac{ \theta (x_i^F)^T x_j^S } { \Vert \theta(x_i^F) \Vert \cdot \Vert x_j^S \Vert  } \right),
\end{equation}
where $f(x_i^F,x_j^S)$ denotes the correlation score between the $i$-th frequency feature vector and the $j$-th spatial feature vector.

Next, we feed forward the spatial feature $x^S$ and perform a matrix multiplication between $x^S$ and the attention score matrix, and then resize it back to the original size $C \times H \times W$. 
In this way, the original spatial features are rescaled taking into account the correlation with frequency features.
Thus, the output mixed feature $y$ can be calculated as:
\begin{equation}
    y_i = \frac{1}{\mathcal{C}(x)} \sum_{\forall j} f(x_i^F,x_j^S) \cdot x_j^S,
\end{equation}
where $i$ is the position index of the output feature. The input spatial feature $x^S_j$ at each position is aggregated into position $i$ with the attention score $f(x_i^F,x_j^S)$. The output feature is normalized by $\mathcal{C}(x)=\sum_{\forall j} f(x_i^F,x_j^S)$.

Compared to the original non-local block, our FSCAB contains fewer parameters by removing unnecessary embedding functions $\phi$ and $g$, being able to fuse features efficiently and effectively.

Such multi-domain feature fusion brings in two advantages:
on one hand, since the frequency and spatial features are extracted in a global and local manner respectively, the mixed features generated by our FSCAB achieve cross-scale aggregation.
Such cross-scale aggregation is beneficial to image inpainting since high-level semantic features can guide the completion of low-level features.
On the other hand, the mixed features can make full use of the characteristics of both the frequency and spatial domains, being able to not only capture the frequency information but also achieve visually pleasing results.

\subsection{Loss Functions}
To effectively guide the training process, it is important to devise the loss function to measure the distance between the generated images and ground truth. 
Therefore, we introduce multiple loss functions from different aspects as follows.
Motivated by PatchGAN~\cite{patchgan}, we adopt a discriminator that works on local levels and discriminates whether a patch is real or fake. 

Let $G$ be the generator and $D$ be the discriminator. 
Given a binary mask $M$ ($1$ for holes) and a ground truth image $I_{gt}$, the input image  with hole $I_{in}$ is formulated as dot product of $I_{gt}$ and $M$, $I_{in} = I_{gt} \odot  M.$ 
Taking the input image $I_{in}$, image inpainting aims to restore high-quality and semantic consistent result $I_{out}=G(I_{in},M)$.

Firstly, we introduce the frequency loss to penalize frequency deviation from the ground truth images and impede loss of frequency information.
In order to stabilize training, we transform $\mathcal{F}$ from the complex number domain to the real number domain and use a log form:
\begin{equation}
\begin{split}
    \mathcal{F}_R(I)(u,v) = \log ( 1 + \sqrt{ \left[ {\rm Real} \mathcal{F}(I)(u,v) \right]^2}  \\
     +\sqrt{ \left[ {\rm Imag} \mathcal{F}(I)(u,v) \right]^2} +\epsilon ),
\end{split}
\end{equation}
where $\rm Real$ and $\rm Imag$ indicate the real and imaginary parts of $\mathcal{F}(I)(u,v)$ and $\epsilon=1 \times 10^{-8}$ is included for numerical stability.
Then we calculate the L1 reconstruction loss between the output of the frequency branch $I_{fre}$ and the ground truth image $I_{gt}$ in the Fourier space:
\begin{equation}
    \mathcal{L}_{fre} = \Vert \mathcal{F}_R(I_{gt}) - \mathcal{F}_R(I_{fre})  \Vert_1 .
\end{equation}

Secondly, we adopt the L1 loss between the inpainted result $I_{gt}$ and the ground truth $I_{out}$ in the spatial domain.
We calculate the loss regarding only the unmasked pixels as:
\begin{equation}
    L_{rec}=\Vert (1-M) \odot (I_{gt} - I_{out}) \Vert_1 ,
\end{equation}
where $\odot$ denotes element-wise multiplication.

Thirdly, we use an adversarial loss to guarantee that our model can generate realistic images.
The adversarial loss is calculated as:

\begin{equation}
    \mathcal{L}_{D} = -{E}_{{I}_{gt}} [\log(D({I}_{gt}))] - {E}_{{I}_{out}} [\log(1 - D({I}_{out}))]
\end{equation}
\begin{equation}
    \mathcal{L}_{G} = -{E}_{I_{out}} [\log(D(I_{out}))]
\end{equation}

What is more, following pix2pixHD~\cite{pix2pixHD}, we adopt the feature matching loss to match intermediate representations of the discriminator as
\begin{equation}
    \mathcal{L}_{fm} =  {E} 
     \left(
     \sum_{i=1}^L \frac{1}{N_i} \Vert D^i(I_{gt}) - D^i(I_{out}) \Vert_1  
     \right),
\end{equation}
where $N_i$ denotes the number of elements in the $i$-th layer, and $L$ is the number of total layers of the discriminator.

Lastly, we adopt a perceptual loss to evaluate the discrepancy between features extracted from the inpainted and ground truth images by a pretrained network $\Phi(\cdot)$.

\begin{equation}
    \mathcal{L}_{perc} = \sum_{i=1}^T \Vert \Phi^i(I_{gt}) - \Phi^i(I_{out}) \Vert_2,
\end{equation}
where $T$ is the total number of layers in the network, and $\Phi^i$ denotes the $i$-th activation map of the pretrained network ResNet50 with dilated convolutions, following LaMa~\cite{lama}. The final loss function for our model is:
\begin{equation}
\mathcal{L} = \mathcal{L}_{fre} + 
            \lambda_r \mathcal{L}_{rec} +
            \lambda_g \mathcal{L}_{G} +
            \lambda_{fm} \mathcal{L}_{fm}+
            \lambda_p \mathcal{L}_{perc}.
\end{equation}
 We set $\lambda_r=10, \lambda_g=10, \lambda_{fm}=100, \lambda_p=30$ empirically to form a linear combination of the above losses.

\section{Experiments}
\subsection{Experiment Setup}
\noindent{\textbf{Datasets.}}
Our model is trained on two public datasets: CelebA-HQ~\cite{celeba-hq} and Places~\cite{place}.
The CelebA-HQ dataset is a high-quality version of CelebA~\cite{celeba} that consists of $30,000$ human face images.
We randomly sample $26,000$ of them as training images, $2000$ as validation images, and $2,000$ as test images.
The Places dataset contains more than $10$ million images, including more than $434$ unique scene categories.
We randomly sample $100,000$ training images, $2,000$ validation images and $30,000$ test images,  

\noindent{\textbf{Evaluation Metrics.}}
All results are evaluated by the metrics of Frechet Inception Distance (FID)~\cite{fid},  Learned Perceptual Image Patch Similarity (LPIPS)~\cite{lpips} and Structural Similarity (SSIM)~\cite{ssim}, all of which are more consistent with human visual neurobiology and perception.
FID compares the distribution of generated images with that of real images, and LPIPS calculates the L2 distance of features extracted by a pretrained model.
SSIM is composed of three measurements with regard to luminance, contrast, and structure.

\begin{figure}[t]
\centering

    \begin{tabular}{ccc}

    \includegraphics[width=0.13\textwidth]{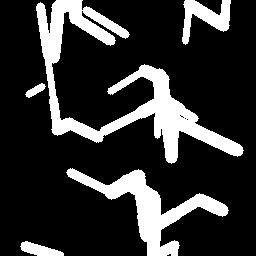}    &
    \includegraphics[width=0.13\textwidth]{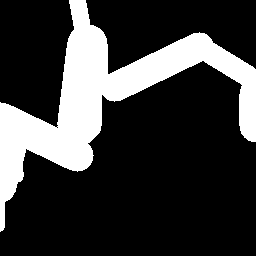}    &
    \includegraphics[width=0.13\textwidth]{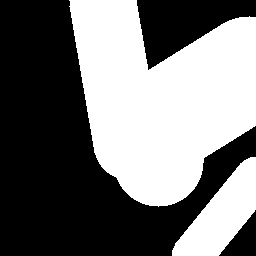}    \\

    (a) Thin &
    (b) Medium &
    (c) Thick
    \\

    \end{tabular}

\vspace{-2mm}
\caption{Samples of different types of masks.}
\vspace{-5mm}
\label{fig:masks}
\end{figure}

\noindent{\textbf{Implementation Details.}}
For the spatial branch, we adopt a UNet~\cite{unet} architecture with $4$ downsampling residual blocks and $4$ upsampling residual blocks.
There are $5$ residual blocks in the frequency branch.
Our model is trained using $256 \times 256$ images with random masks generated following the settings in LaMa~\cite{lama}.
We use Adam optimizer~\cite{adam}  with $\beta_1 = 0.9$ and $\beta_2 = 0.999$, and set fixed learning rates as $0.001$ and $0.0001$ for the generator and discriminators respectively. 
Our model is trained on NVIDIA V100 GPUs with a batch size of $30$ for $1$ million iterations.
We prepare test sets with irregular random masks of different widths (thin, medium, thick) following LaMa~\cite{lama}.
This strategy uniformly uses samples from polygon chains extended by rectangles of high random width and arbitrary aspect ratio.
We set different values of the following parameters to generate different types of masks: a probability of a polygonal chain mask, min number of segments, max number of segments, max length of a segment in polygonal chain, max width of a segment in polygonal chain, min bound for the number of box primitives, max bound for the number of box primitives, min length of a box side, max length of a box side, etc. 
We show samples of masks in Fig.~\ref{fig:masks}.

\subsection{Ablation Study}
To study the effects of individual components in the proposed model FSCN, we conduct several ablation studies on the dataset of CelebA-HQ.
Specifically, we train five model variants including (a) removing the frequency branch and retaining only the spatial branch (``w/o fre branch''), (b) retaining both the frequency and spatial branches but removing the frequency loss (``w/o fre loss''), (c) replacing our feature fusion module FSCAB by concatenating multi-domain features (``concat''), (d) concatenating the frequency and spatial features and then adopt self-attention to fuse multi-domain features (``self-attention''), (e) our final model with both the frequency and spatial branches, the frequency loss and feature fusion block FSCAB.

\begin{table}[!t]
\renewcommand\arraystretch{1.3}
\huge

\center
\begin{center}
\caption{Quantitative evaluation of ablation studies on CelebA-HQ dataset. Our full model equipped with the proposed frequency branch, the frequency loss and feature fusion block FSCAB achieves the best results.}
\label{tab:ablation}
\vspace{-1mm}

\resizebox{1.0\columnwidth}{!}
{
\begin{tabular}{l |cc |cc |cc}
\hline
\multirow{2}{*}{Model}  &  \multicolumn{2}{c|}{Thin} &  \multicolumn{2}{c|}{Medium} &  \multicolumn{2}{c}{Thick} \\
\cline{2-7} 
& FID $\downarrow$ & LPIPS $\downarrow$ &FID $\downarrow$ & LPIPS $\downarrow$ &FID $\downarrow$ & LPIPS $\downarrow$ \\
\hline
w/o fre branch &6.2787 &0.0909 &5.5051 & 0.0826 &5.6125 &0.0948 \\ 
w/o fre loss &5.3982 &0.0793 &5.1040 &0.0770 & 5.3505 &0.0898 \\ \hline
concat &5.4659 &0.0852 &5.0711 &0.0811 &5.3650 &0.0945 \\
self-attention &5.4459 &0.0847 &5.0489 &0.0782 &5.2837 &0.0910 \\ \hline
ours &\textbf{5.1956} &\textbf{0.0768} &\textbf{5.0189} &\textbf{0.0762} &\textbf{5.2123} &\textbf{0.0897}\\ \hline       

\end{tabular}
}
\end{center}
\vspace{-4mm}
\end{table}

\begin{figure}[t]
\centering
    

    \begin{tabular}{cccc}
    \includegraphics[width=0.15\textwidth]{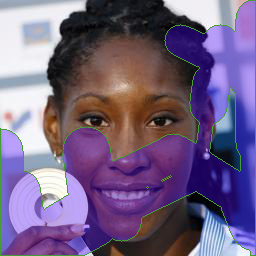}  & \hspace{-6mm}
    \includegraphics[width=0.15\textwidth]{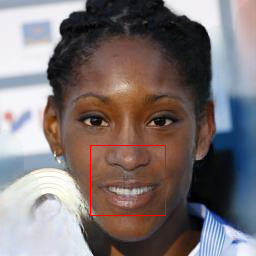} & \hspace{-6mm}
    \includegraphics[width=0.15\textwidth]{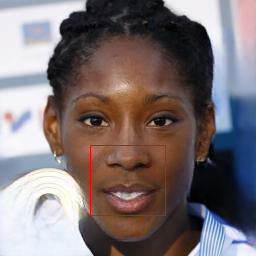}  & \hspace{-6mm}\\
    (a) original &
    (b) w/o fre branch &
    (c) w/o fre loss  & \\
    
    \includegraphics[width=0.15\textwidth]{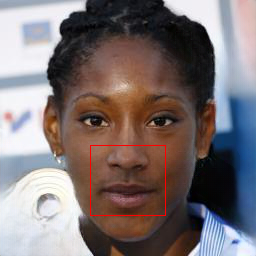}  & \hspace{-6mm}
    \includegraphics[width=0.15\textwidth]{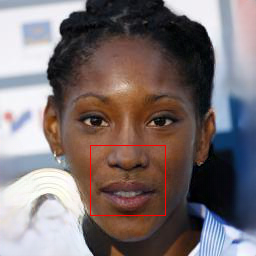}& \hspace{-6mm}
    \includegraphics[width=0.15\textwidth]{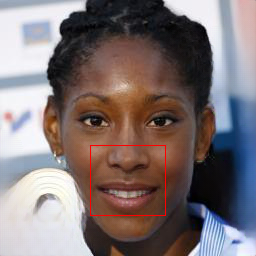}  &
    \\
    
    (d) concat &
    (e) self-attention&
    (f) ours
    
    \end{tabular}



    

\vspace{-2mm}
\caption{Visual ablation study results on CelebA-HQ dataset (zoom in for a better view). The proposed frequency branch, frequency loss and feature fusion block contribute to suppress artifacts and recover high-quality results.}
\vspace{-6mm}
\label{fig:ablation}
\end{figure}

The results of the five model variants are reported in Table~\ref{tab:ablation}. As it is shown, removing the frequency branch results in a significant performance decrease.
The frequency branch can help preserve the frequency information and decrease the discrepancy between the generated images and the ground truth images in the frequency domain, which can be reflected in the spatial space as perceptible details.
Removing the frequency loss causes a further decrease since the frequency loss penalizes deviating from the frequency spectra of the ground truth images and helps ensure frequency consistency.
Visual evaluation in Fig.~\ref{fig:ablation} (b) and (c) is consistent with quantitative experiments, suggesting that the model fails to recover fine-grained details without the frequency branch and frequency loss, e.g. unreal texture fog in (b) and unreasonable color distortion in (c).
Simply concatenating the frequency and spatial features is not able to make full use of multi-domain feature aggregation, and adding self-attention does not help much.
Although image details are restored, such insufficient fusion leads to mismatches in frequency and spatial domains, as shown by the unreasonable color around the mouth and nose in Fig.~\ref{fig:ablation} (d) and the ghost artifacts on the contour of the face in Fig.~\ref{fig:ablation} (e). 
With our FSCAB, the effectively mixed features can take full advantage of the characteristics of both the frequency and spatial features to capture frequency information and obtain visually pleasing results simultaneously.

\subsection{Quantitative Evaluation}
We compare our model with other state-of-the-art methods: LaMa~\cite{lama}, Wavefill~\cite{wavefill}, MAT~\cite{li2022mat}, MADF~\cite{madf}, DMFN~\cite{dmfn}, Deepfillv2~\cite{deepfillv2} and EdgeConnect~\cite{edgeconnect}.
To make the comparison fair, only the publicly available pretrained models are used to compute the evaluation metric values. 
For each dataset, we validate the performance across three different types of masks.

All the results of compared methods are reported in Table~\ref{tab:celeb} for the CelebA-HQ dataset and Table~\ref{tab:place} for the Places dataset. Note that, we abbreviate EdgeConnect as EC and Deepfillv2 as GC. The results demonstrate that our model consistently achieves the best performance on different benchmarks in most cases, no matter in the scenery of a large mask or a thin mask.
On the CelebA-HQ dataset, compared with the previous SOTA method LaMa with only the spatial domain features, FID of our proposed FSCN is improved by $24.02\%$. Compared to previous SOTA method Wavefill with only features in the frequency domain, SSIM is improved from $0.9207$ to $0.9247$. 
Although a thick mask leads to incoherent structure, there is still a significant improvement in all the metrics compared with previous works.
It implies that our model can make full use of informative features from both the frequency and spatial domains, which attributes to our proposed frequency branch, frequency loss, and multi-domain feature fusion block.
Subsequently, our model can capture global semantic information and preserve the overall structure and outline of images, which is beneficial to high-fidelity image inpainting. 

\begin{table*}[t]
\renewcommand\arraystretch{1.5}
\center
\definecolor{darkgreen}{RGB}{40,240,123}
\begin{center}
\caption{
Quantitative evaluation on CelebA-HQ dataset~\cite{celeba-hq}. 
The \textbf{best} and \underline{second best} results are highlighted. }
\label{tab:celeb}
\vspace{-2mm}
\resizebox{0.9\textwidth}{!}
{
\begin{tabular}{c |ccc |ccc |ccc}
\hline
\multirow{2}*{Method}  
& \multicolumn{3}{c|}{Thin} & \multicolumn{3}{c|}{Medium} &  \multicolumn{3}{c}{Thick}   \\
\cline{2-10} 
 & FID $\downarrow$ & LPIPS $\downarrow$ & SSIM $\uparrow$ &FID $\downarrow$ & LPIPS $\downarrow$ & SSIM $\uparrow$  &FID $\downarrow$ & LPIPS $\downarrow$ & SSIM $\uparrow$  \\
\hline

EC $(ICCV'2019)$ \cite{edgeconnect}  &7.0074 &0.0921 &0.9124 &6.1678 &0.0919 &0.9033 &6.9193 &0.1107 &0.8783  \\\hline  

GC $(ICCV'2019)$ \cite{deepfillv2}  &11.1879 &0.1295 &0.8964 &8.1926 &0.1063 &0.8954 &9.9133 &0.1213 &0.8690 \\\hline

MADF $(TIP'2021)$ \cite{madf}  &6.9000 &0.0858 &0.9166 &10.2428 &0.1027 &0.8992 &16.4815 &0.1331 &0.8706 \\ \hline

DMFN $(arXiv'2020)$ \cite{dmfn}  &7.8631 &0.1055 &0.9070 &6.2192 &0.0888 &0.9062 &6.9401 &0.1041 &0.8839 \\ \hline

Wavefill $(ICCV'2021)$ \cite{wavefill}   &\underline{5.2502} &\underline{0.0790} &0.9207 &5.4269 &0.0815 &0.9096 &5.6187 &0.0963 &0.8889 \\ \hline

LaMa $(WACV'2022)$ \cite{lama}  &5.7965 &0.0825 &\underline{0.9208} &\underline{5.2425} &\underline{0.0792} &\underline{0.9141} &\underline{5.5395} &\underline{0.0924} &\underline{0.8942} \\\hline

Our FSCN &\textbf{4.6739} &\textbf{0.0762} &\textbf{0.9247} &\textbf{4.9603} &\textbf{0.0766} &\textbf{0.9162} &\textbf{5.4149} &\textbf{0.0897} &\textbf{0.8971}
\\ \hline
\end{tabular}
}
\vspace{-3mm}
\end{center}
\end{table*}

\definecolor{darkgreen}{RGB}{40,240,123}

\begin{table*}[!t]
\renewcommand\arraystretch{1.5}
\center
\begin{center}
\caption{
Quantitative evaluation on Places dataset~\cite{place}.
The \textbf{best} and \underline{second best} results are highlighted.}
\label{tab:place}
\vspace{-2mm}

\resizebox{0.9\textwidth}{!}{
\begin{tabular}{c|ccc|ccc|ccc }
\hline
\multirow{2}*{Method}  
&\multicolumn{3}{c|}{Thin} & \multicolumn{3}{c|}{Medium} & \multicolumn{3}{c}{Thick} \\
\cline{2-10} 
 & FID $\downarrow$ & LPIPS $\downarrow$ & SSIM $\uparrow$    &FID $\downarrow$ & LPIPS $\downarrow$ & SSIM $\uparrow$  &FID $\downarrow$ & LPIPS $\downarrow$ & SSIM $\uparrow$  \\
\hline




  


EC (ICCV'2019) \cite{edgeconnect} &1.3542 &0.1110 &\textbf{0.9347} &3.6796 &0.1347 &0.8720 &8.5293 &0.1588 &0.8492 \\\hline 

GC (ICCV'2019) \cite{deepfillv2} &1.0668 &0.1044 &0.9061 &2.7125 &0.1300 &0.8688 &5.2606 &0.1541 &0.8391 \\ \hline

Wavefill (ICCV'2021) \cite{wavefill} &0.9794 &0.0994 &0.9075 &\underline{1.1959} &0.2828 &0.6473 &1.3204 &0.3679 &0.5467 \\ \hline

MADF (TIP'2021) \cite{madf} & \underline{0.5787} &\underline{0.0857} &0.9140 &1.6844 &0.1145 &0.8796 &3.7856 &0.1377 &0.8560 \\ \hline
  
MAT (CVPR'2022) \cite{li2022mat} &0.6298 &0.0951 &0.9033 &1.2386 &0.1216 &0.8676 &\textbf{1.8765} &0.1430 &0.8428 \\ \hline

LaMa (WACV'2022) \cite{lama} &0.6380 &0.0905 &0.9153 &1.3152 &\underline{0.1121} &\underline{0.8857} &2.2150 &\underline{0.1334} &\textbf{0.8669} \\ \hline
  
Our FSCN &\textbf{0.5436} &\textbf{0.0807} &\underline{0.9184} &\textbf{1.1555} &\textbf{0.1077} &\textbf{0.8860} &\underline{2.1703} &\textbf{0.1316} &\underline{0.8640}\\ \hline
\end{tabular}
}
\end{center}

\vspace{-4mm}
\end{table*}

On the Places dataset, our results still outperform most of the other methods, both in the frequency domain and spatial domain. In the case of a thick mask, it is slightly worse than two competitive methods LaMa and MAT. One possible reason is that thick masks pose higher requirements on the generation ability for models to recover high-fidelity results. Both LaMa and MAT are complex with a larger amount of calculation and huge model capacity. Specifically, LaMa uses approximately $ 2\times$ and MAT employs about $5\times$ parameters than FSCN.
However, these two methods are not superior in terms of visualization results in Fig.~\ref{fig:place}.
In terms of EdgeConnect under a thin mask, it uses edge map as prior knowledge to guide the reconstruction. Thus it helps preserve structural information and obtains better SSIM values.

\begin{figure}[t]


\begin{tabular}{cc}
\includegraphics[width=0.42\columnwidth]{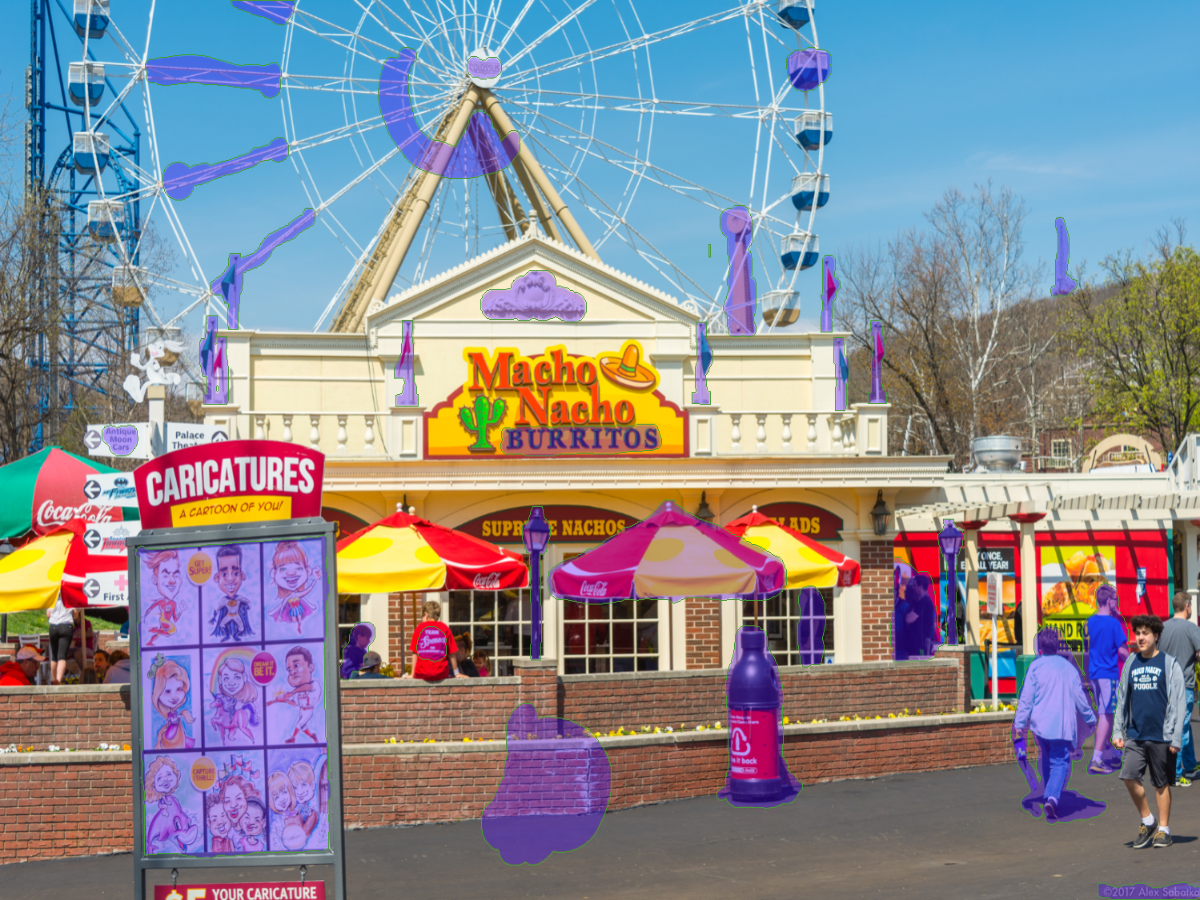} & \hspace{-1mm}
\includegraphics[width=0.42\columnwidth]{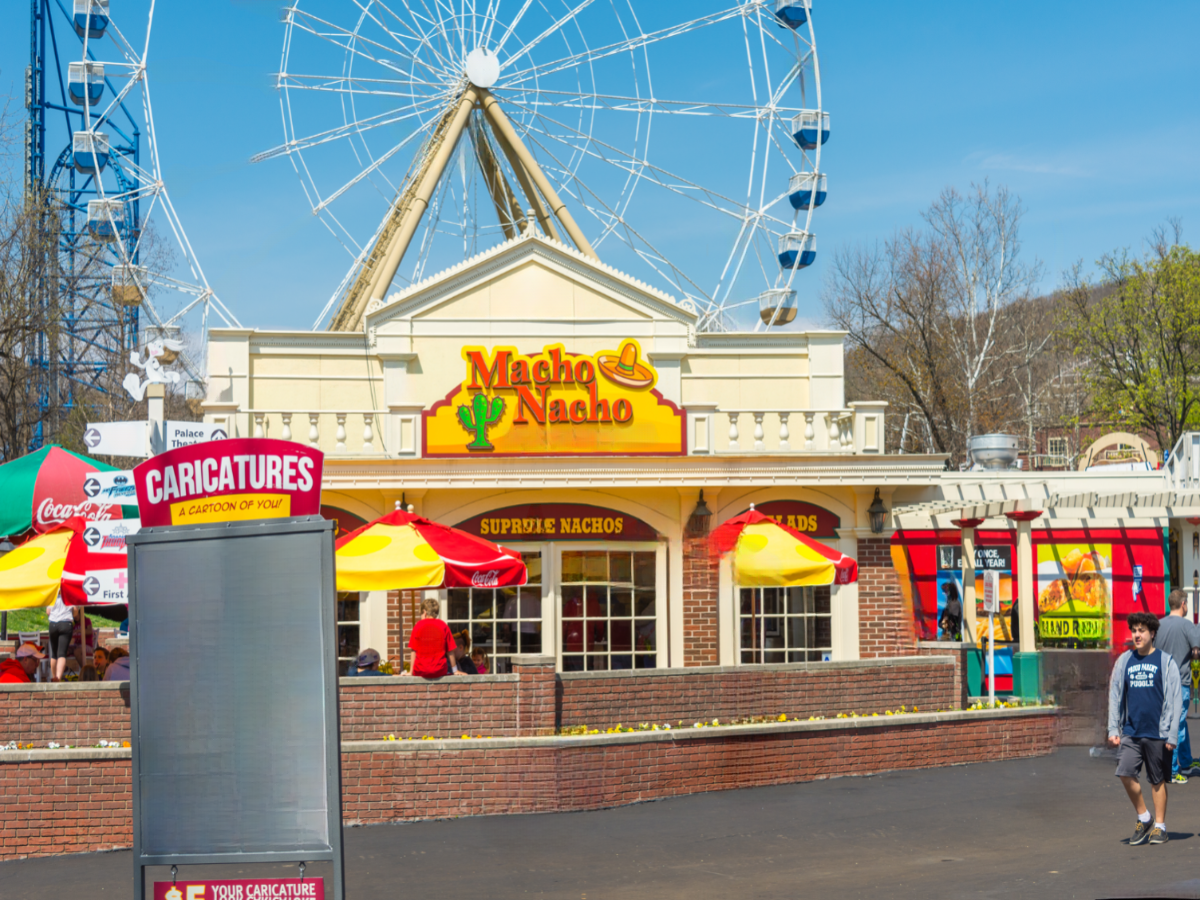}  \\
\end{tabular}


\begin{tabular}{cc}
\includegraphics[width=0.42\columnwidth]{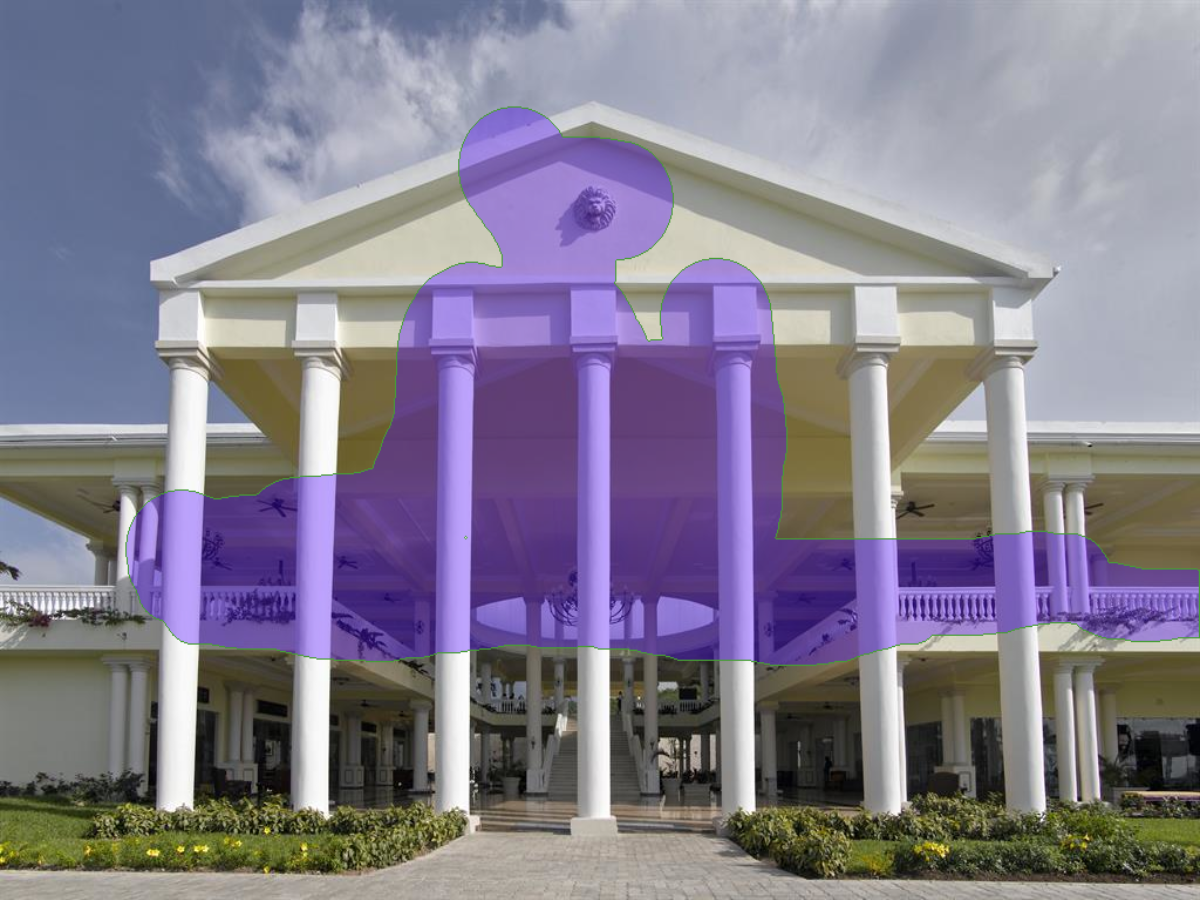} & \hspace{-1mm}
\includegraphics[width=0.42\columnwidth]{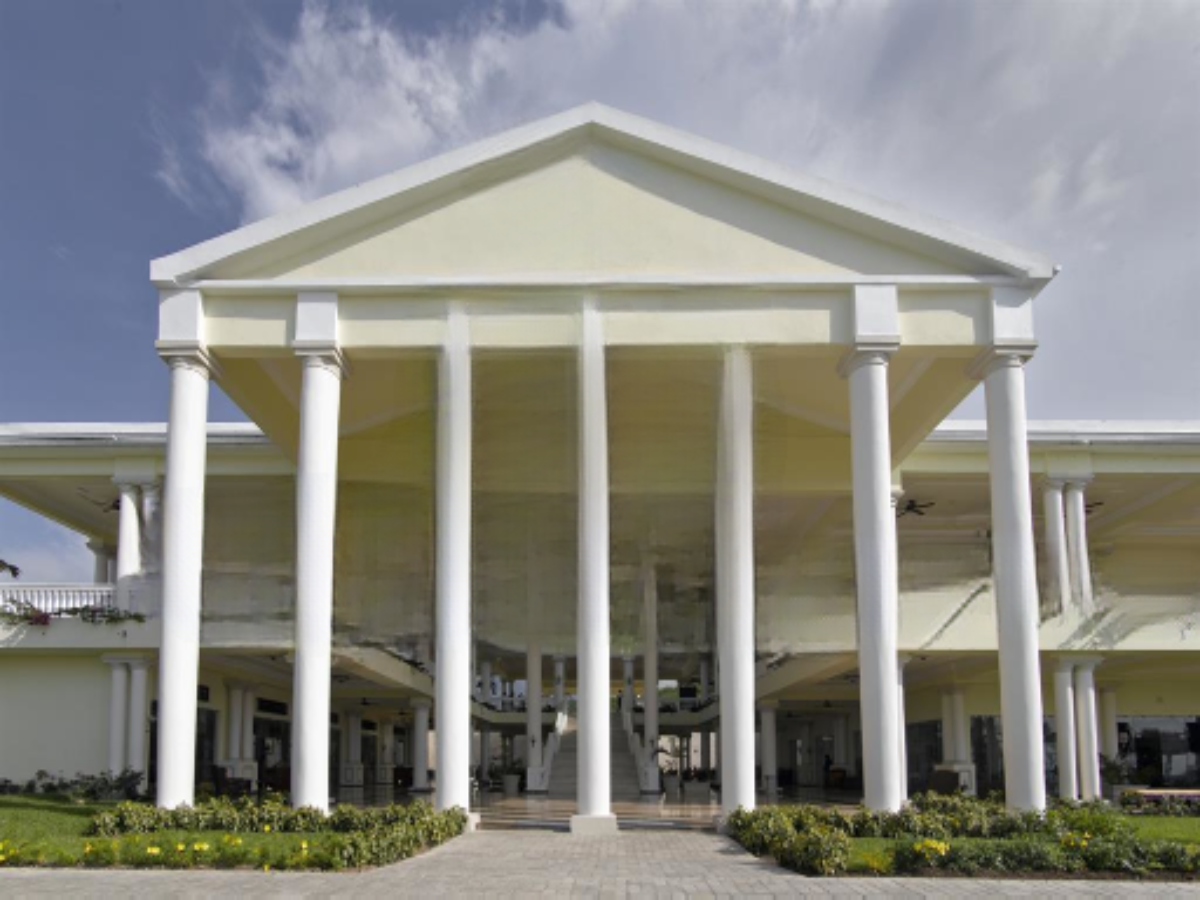}  \\

(a) Masked Input  & \hspace{-4mm}
(b) Output
\end{tabular}


\vspace{-2mm}
\caption{ Our model produces realistic results for the real-world object removal task (zoom in for a better view).}
\vspace{-5mm}
\label{fig:object}
\end{figure}

\subsection{Qualitative Evaluation}
\begin{figure*}[htb!]
\centering
    \begin{tabular}{ccccccc}
    \includegraphics[width=0.16\textwidth]{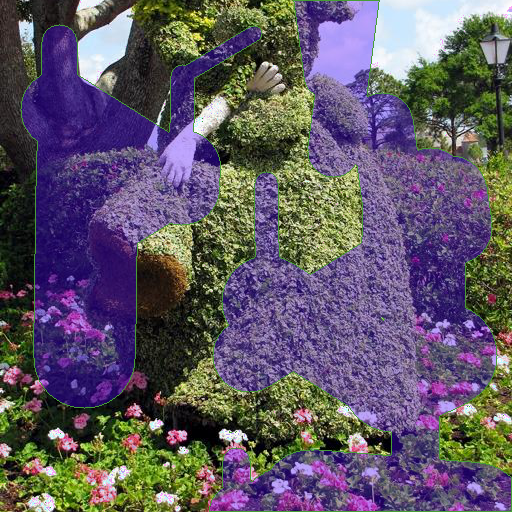}    & \hspace{-4mm}
    \includegraphics[width=0.16\textwidth]{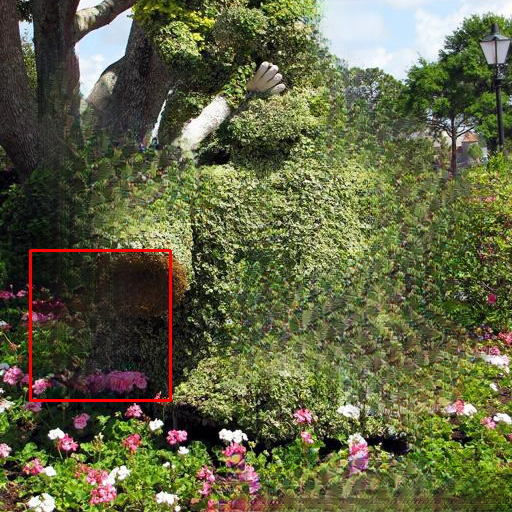}    & \hspace{-4mm}
    \includegraphics[width=0.16\textwidth]{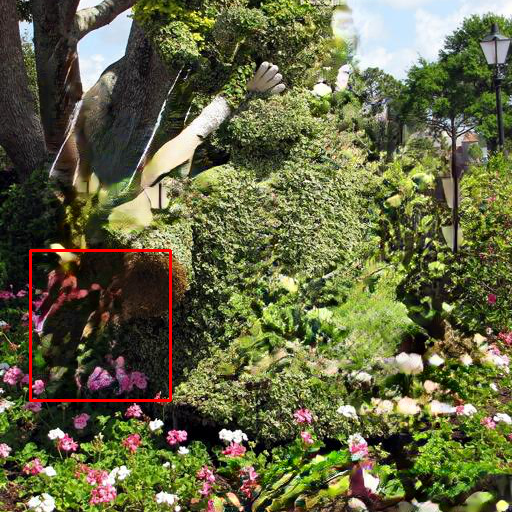}    & \hspace{-4mm}
    \includegraphics[width=0.16\textwidth]{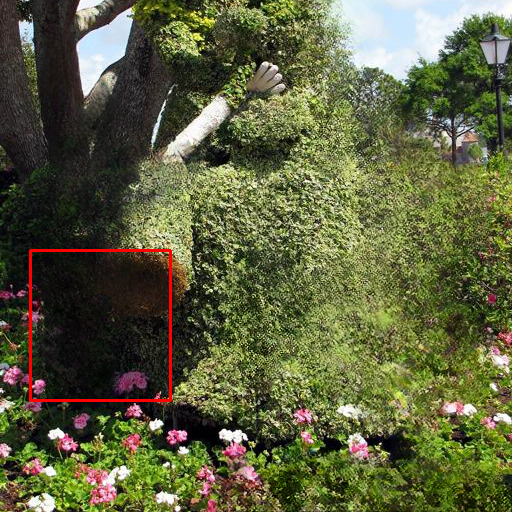}    & \hspace{-4mm}
    \includegraphics[width=0.16\textwidth]{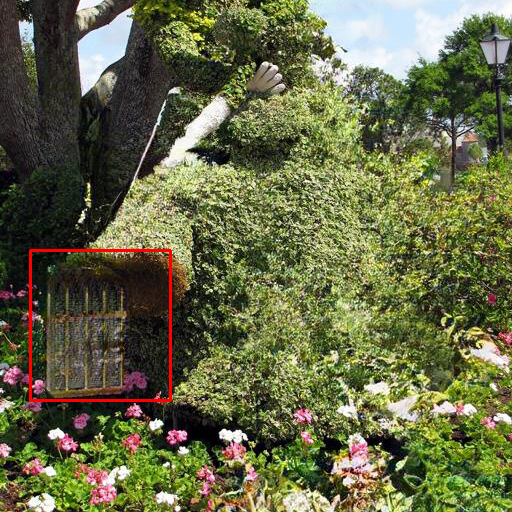}   & \hspace{-4mm}
    \includegraphics[width=0.16\textwidth]{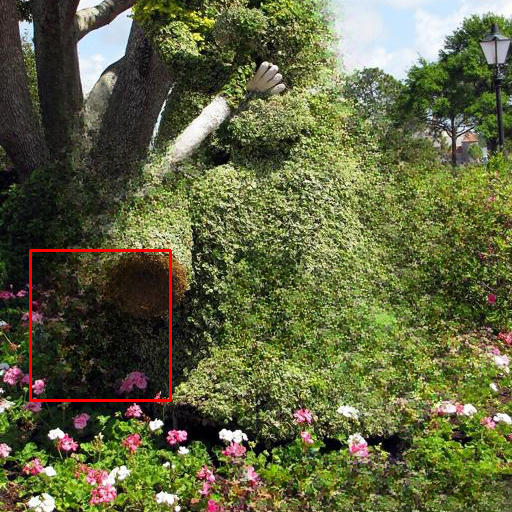}  
    \\
    \end{tabular}
    \begin{tabular}{ccccccc}
    \includegraphics[width=0.16\textwidth]{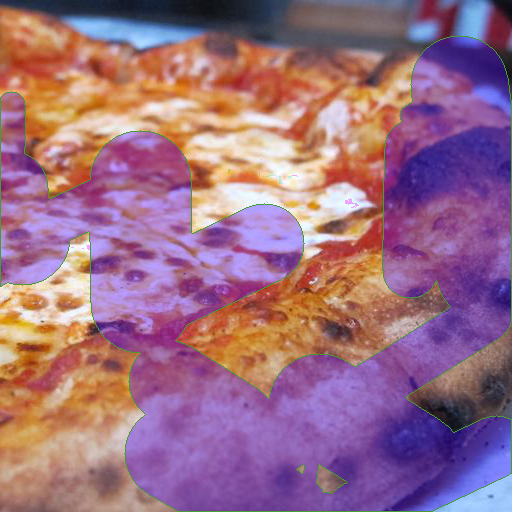}    & \hspace{-4mm}
    \includegraphics[width=0.16\textwidth]{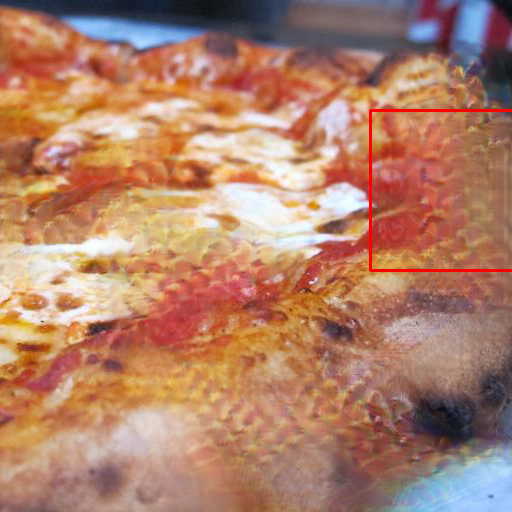}    & \hspace{-4mm}
    \includegraphics[width=0.16\textwidth]{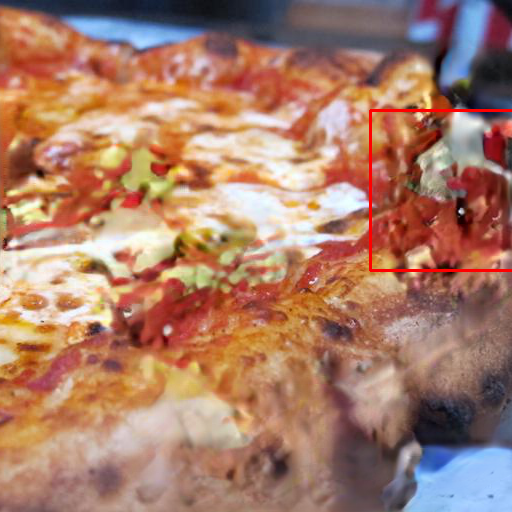}    & \hspace{-4mm}
    \includegraphics[width=0.16\textwidth]{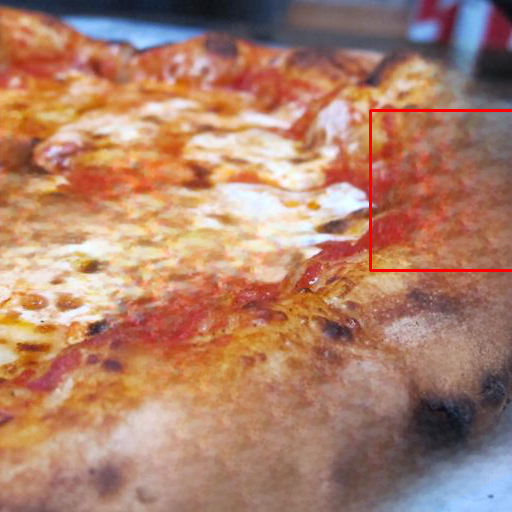}    & \hspace{-4mm}
    \includegraphics[width=0.16\textwidth]{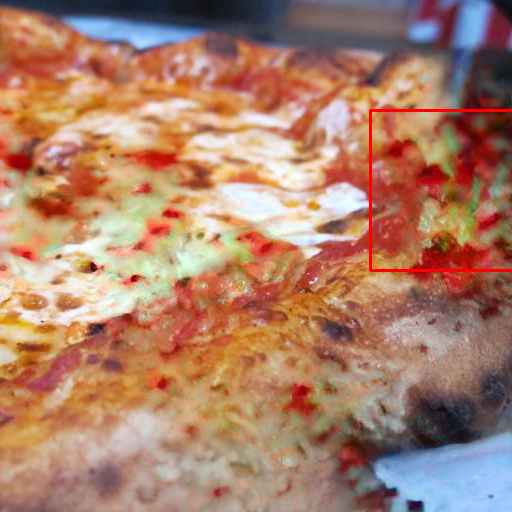}     & \hspace{-4mm}
    \includegraphics[width=0.16\textwidth]{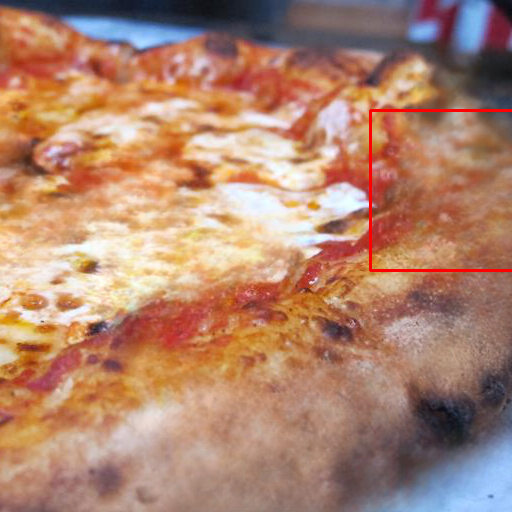}   
    \\

    (a) Original  &\hspace{-4mm}
    (b) EdgeConnect  &\hspace{-4mm}
    (c) Deepfillv2  &\hspace{-4mm}
    (d) LaMa &\hspace{-4mm}
    (e) MAT &\hspace{-4mm}
    (f) Ours
    \\

    \end{tabular}

\vspace{-2mm}
\caption{Qualitative comparison of the SOTA methods with our model on the Places~\cite{place} dataset. Our results are more visually pleasing compared with SOTA methods. Zoom in for a better view.}
\vspace{-3mm}
\label{fig:place}
\end{figure*}

\begin{figure*}[t]
\centering
    \begin{tabular}{ccccccc}
    \includegraphics[width=0.16\textwidth]{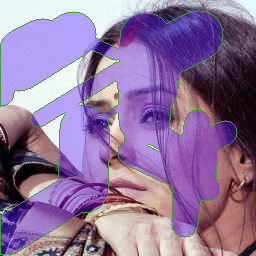}    & \hspace{-4mm}
    \includegraphics[width=0.16\textwidth]{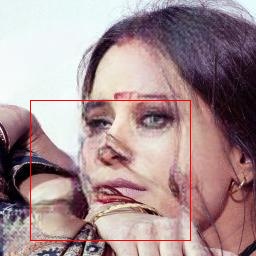}   & \hspace{-4mm}
    \includegraphics[width=0.16\textwidth]{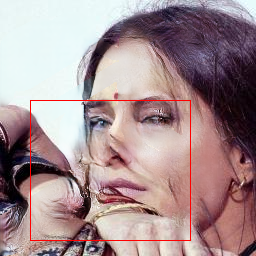}    & \hspace{-4mm}
    \includegraphics[width=0.16\textwidth]{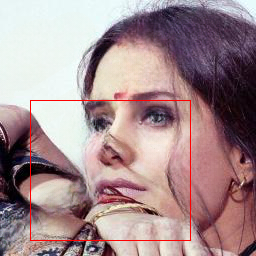}    & \hspace{-4mm}
    \includegraphics[width=0.16\textwidth]{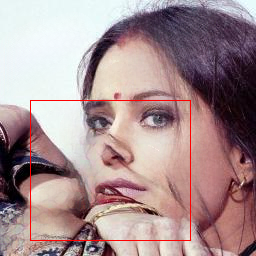}    & \hspace{-4mm}
    \includegraphics[width=0.16\textwidth]{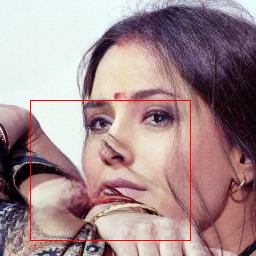}  
    \\
    \end{tabular}

    \begin{tabular}{ccccccc}
    \includegraphics[width=0.16\textwidth]{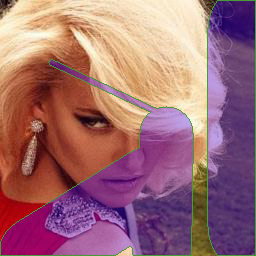}    & \hspace{-4mm}
    \includegraphics[width=0.16\textwidth]{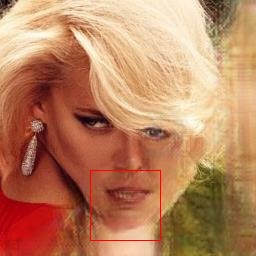}   & \hspace{-4mm}
    \includegraphics[width=0.16\textwidth]{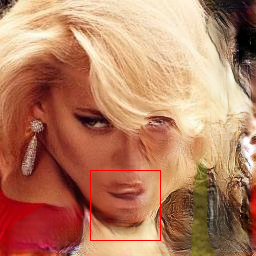}    & \hspace{-4mm}
    \includegraphics[width=0.16\textwidth]{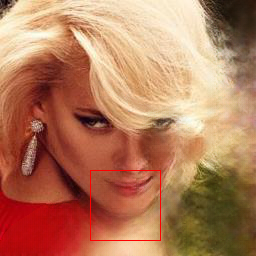}    & \hspace{-4mm}
    \includegraphics[width=0.16\textwidth]{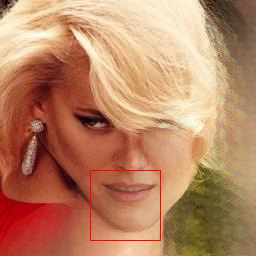}    & \hspace{-4mm}
    \includegraphics[width=0.16\textwidth]{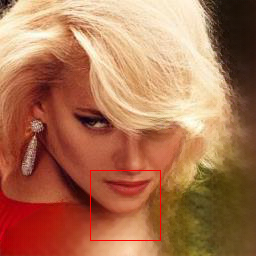}  \\
    (a) Original  &\hspace{-4mm}
    (b) EdgeConnect &\hspace{-4mm}
    (c) Deepfillv2  &\hspace{-4mm}
    (d) LaMa &\hspace{-4mm}
    (e) WaveFill  &\hspace{-4mm}
    (f) Ours
    
    \\
    \end{tabular}
    
\vspace{-2mm}
\caption{Qualitative comparison of the SOTA methods with our model on the CelebA-HQ~\cite{celeba-hq} dataset. Our model generates more realistic inpainting results with fewer artifacts compared with SOTA methods. Zoom in for a better view.}
\vspace{-4mm}
\label{fig:celeb}
\end{figure*}

As shown in Fig.~\ref{fig:place} and Fig.~\ref{fig:celeb}, we also compare exemplar visual results of various existing SOTA inpainting methods with ours.
The figures show that our model produces visually pleasing results with finer details and image structures, outperforming other methods significantly.
For example, in the ``pizza'' image in bottom row of Fig.~\ref{fig:place}, MAT recovers images with abnormal color and LaMa suffers from grid-like artifacts, while our model produces more faithful results with fewer artifacts and finer fine-grained details. What is more, the model complexity also brings side effects to MAT. As shown in Fig.~\ref{fig:place}, MAT generates a semantically inconsistent object ``windows'' in the bush (first row), which does not match the image context. Since deep neural networks pay more attention to low-frequency components, the generated frequency spectra often deviate from the ground truth spectra. This can be reflected in the visual space as the lack of important fine-grained details, leading to unpleasant perceptual quality. However, the frequency branch in FSCN helps prevent the deviation of important frequency components due to the inherent bias of neural networks and decreases the distance between the generated images and the ground truth images in the frequency domain. 
Also, in Fig.~\ref{fig:celeb}, FSCN recovers reasonable face shape and edge details, while LaMa fails in recovering fine-grained details of the facial region, and WaveFill leads to unreal facial organs. Since the frequency and spatial features are extracted in a global and local manner respectively, the well-fused cross-scale features are beneficial to the image inpainting task since high-level semantic features can guide the completion of low-level features, resulting in better perceptual quality. See supplementary materials for more results.


\subsection{Real-world Applications}
We also conduct experiments on real-world applications as shown in Fig.~\ref{fig:object}. More results can be seen in supplementary materials. 
In the first case of object removal, with dense structures and textures, our model recovers semantically consistent textures. In the second case, it effectively restores the building structure and recovers the background with fine-grained details, producing a visually realistic image, which demonstrates the effectiveness of our model. 

\subsection{Model Complexity Analyses}

\begin{figure}[t]
\centering
\includegraphics[width=0.9\columnwidth]{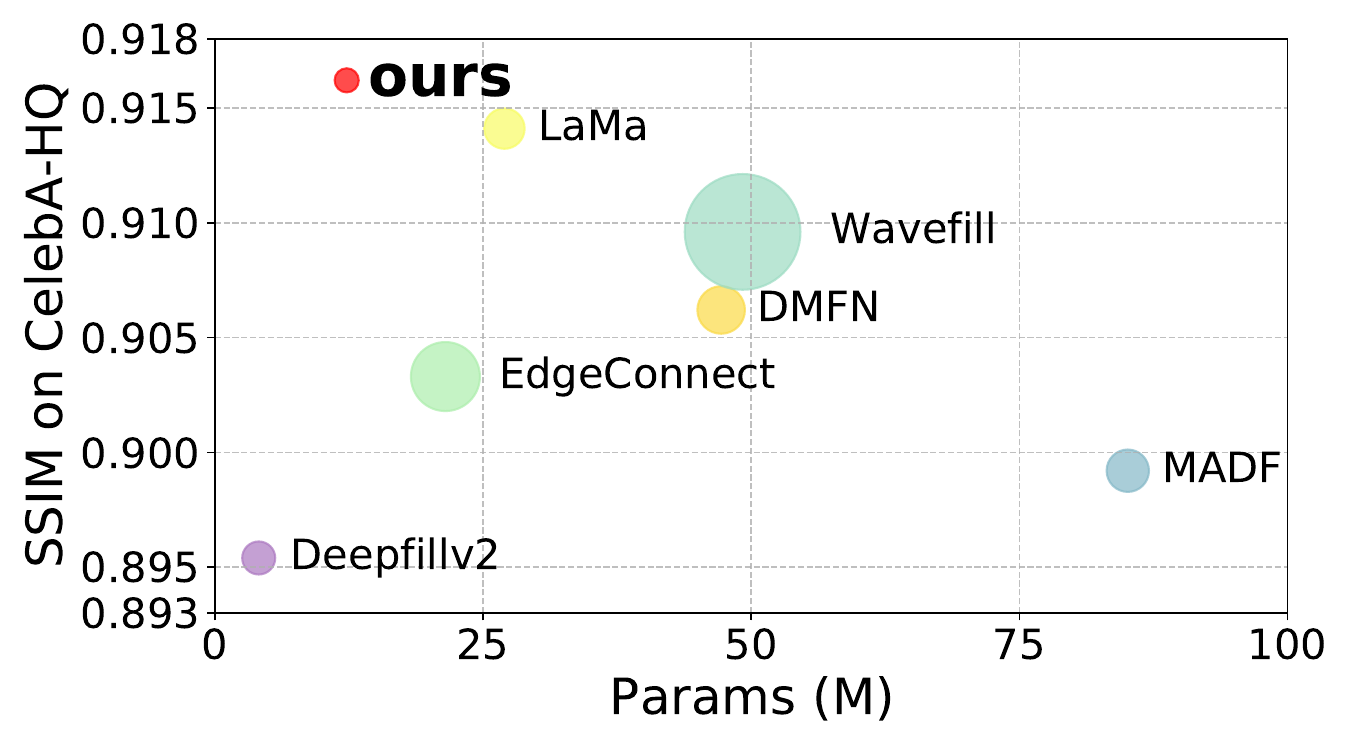} 
\vspace{-2mm}
\caption{Model size comparison. 
The smaller the circle, the fewer the MACs.} 
\vspace{-5mm}
\label{fig:params}
\end{figure}

We compare model parameters and MACs with other advanced inpainting methods in Fig.~\ref{fig:params}. It is observed that our model is located in the upper left, indicating that our model can achieve better results with fewer parameters and less computation cost. 
Our model is superior to all other methods with fewer parameters and MACs. More specifically, our model surpasses LaMa with regard to all evaluation metrics on the CelebA-HQ dataset with only half parameters ($27.0$M versus $12.3$M) and one-third MACs ($42.85$G versus $15.68$G). It illustrates the superiority of our model over other state-of-the-art approaches.

\section{Conclusion}
In this paper, we propose a Frequency-Spatial Complementary Network (FSCN) for high-fidelity image inpainting.
The network contains an extra Frequency Branch and Frequency Loss to exploit features in the frequency domain and ensure frequency consistency, and a Frequency-Spatial Cross-Attention Block to fuse multi-domain features for better completion. 
Quantitative and qualitative experiments demonstrate that FSCN can effectively produce high-fidelity results with semantically consistent structures and fine-grained details, outperforming the SOTA methods with fewer parameters and less computational cost.


{\small
\bibliographystyle{ieee_fullname}
\bibliography{myreferences}
}

\end{document}